\documentclass[lettersize,journal]{IEEEtran}
\usepackage[implicit=false]{hyperref}
\usepackage[draft]{changes} % 草稿模式显示所有更改
\definechangesauthor[name={Your Name}, color=blue]{user}
\usepackage{amsmath,amsfonts}
\usepackage{algorithmic}
\usepackage{array}
\usepackage[caption=false,font=normalsize,labelfont=sf,textfont=sf]{subfig}
\usepackage{textcomp}
\usepackage{stfloats}
\usepackage{url}
\usepackage{verbatim}
\usepackage{graphicx}
\def\BibTeX{{\rm B\kern-.05em{\sc i\kern-.025em b}\kern-.08em
    T\kern-.1667em\lower.7ex\hbox{E}\kern-.125emX}}
\usepackage{balance}
\usepackage{multirow}
\usepackage{makecell}
\usepackage{tabularx}
\usepackage[labelformat=simple]{subcaption}
\usepackage{tikz,xcolor}
\hypersetup{hidelinks,
	colorlinks=true,
	allcolors=black,
	pdfstartview=Fit,
	breaklinks=true}

\definecolor{lime}{HTML}{A6CE39}
\DeclareRobustCommand{\orcidicon}{
	\begin{tikzpicture}
		\draw[lime, fill=lime] (0,0)
		circle[radius=0.16]
		node[white]{{\fontfamily{qag}\selectfont \tiny \.{I}D}};
	\end{tikzpicture}
	\hspace{-2mm}
}
\foreach \x in {A, ..., Z}{%
	\expandafter\xdef\csname orcid\x\endcsname{\noexpand\href{https://orcid.org/\csname orcidauthor\x\endcsname}{\noexpand\orcidicon}}
}

\begin{document}
	
\title{Multimodal-Aware Fusion Network for Referring Remote Sensing Image Segmentation}
\author{Leideng Shi\orcidA{}, Juan Zhang\orcidB{}
\thanks{Date of current version 24 September
2024. This work was supported in part by Shanghai Local Capacity Enhancement project (No. 21010501500), in part by "Science and Technology Innovation Action Plan" of Shanghai Science and Technology Commission for social development project under Grant 21DZ1204900.(Corresponding author: Juan Zhang.)

Leideng Shi and Juan Zhang are with the School of Electronic and Electrical Engineering, Shanghai University of Engineering Science, Shanghai 201600, China (e-mail:shileideng0121@gmail.com; zhang-j@foxmail.com)
}}

\markboth{IEEE GEOSCIENCE AND REMOTE SENSING LETTERS}%
{How to Use the IEEEtran \LaTeX \ Templates}

\maketitle

\begin{abstract}
Referring remote sensing image segmentation (RRSIS) is a novel visual task in remote sensing images segmentation, which aims to segment objects based on a given text description, with great significance in practical application. Previous studies fuse visual and linguistic modalities by explicit feature interaction, which fail to effectively excavate useful multimodal information from dual-branch encoder. In this letter, we design a multimodal-aware fusion network (MAFN) to achieve fine-grained alignment and fusion between the two modalities. We propose a correlation fusion module (CFM) to enhance multi-scale visual features by introducing adaptively noise in transformer, and integrate cross-modal aware features. In addition, MAFN employs multi-scale refinement convolution (MSRC) to adapt to the various orientations of objects at different scales to boost their representation ability to enhances segmentation accuracy. Extensive experiments have shown that MAFN is significantly more effective than the state of the art on RRSIS-D datasets. The source code is available at https://github.com/Roaxy/MAFN.
\end{abstract}


\section{Introduction}
\IEEEPARstart{I}{n} recent years, referring remote sensing image segmentation (RRSIS), as one of the most critical tasks in the field of remote sensing, has experienced impressive advancements. RRSIS \cite{rrsis} allows us to monitor climate changes, manage natural disasters, and plan urban infrastructure development efficiently. Unlike traditional remote sensing image segmentation, RRSIS possesses a complex understanding of spatial information and geographic nuances, enabling it to segment specific objects in remote sensing images as specified by textual descriptions. The targets of segmentation are usually indicated by arbitrarily composed expression, which can consist of different entities, positions, actions, and attributes. Therefore, the key challenge in RRSIS is to generate specific visual features from the given text to guide remote sensing image segmentation.

As the forefront of remote sensing image segmentation, RRSIS is increasingly dedicating efforts towards extracting visual features and combining them with natural language features. With the rapid development of Transformer, cross-modal fusion based Transformer methods gradually entered the mainstream. In visual grounding domain, Deng et al. \cite{transvg} introduced a transformer-based framework, which jointly input visual and language tokens from two branches into the Transformer for relational reasoning. Zhan et al. \cite{rsvg} applied this method to remote sensing images with complicated scales to excavate the potential relationship between text semantics and visual perception. Although effective, cross-modal interactions occur only after feature encoding, and previous approaches fail to effectively leverage the Transformer in the encoder to extract helpful multimodal context \cite{lavt}. Recent RRSIS methods, such as RMSIN \cite{RMSIN}, abandoned the complex cross-modal decoder and employed dual-branch structures to jointly embed linguistic and visual features during visual encoding process. Although this approach enhances visual priors and facilitates cross-modal fusion, it exploits explicit feature interactions to fuse dual-branch information, often ignoring complex correlations between modalities, which makes it difficult to learn fine-grained visual and linguistic alignments.

To address the problem, we propose the multimodal-aware fusion network (MAFN) that enriches the fusion of visual information and language features through the use of location correlation perception. We consider two main issues in our method. On the one hand, we design a correlation fusion module (CFM) for text-driven feature and visual fine-grained alignment. In detail, we improve Swin Transformer layer by adding adaptive noise to enhance modality learning capability \cite{swin}, called AN-SwinT. To effectively excavate correspondences between branches and further enhance the feature representation, CFM can make full use of the correlation-aware design to establish position correspondence of the text embeddings with the related spatial visual features. On the other hand, For refining the segmentation of object edges in remote sensing images, we present multi-scale refinement convolution (MSRC), which can refine edge information and improve segmentation accuracy. Therefore, based on these modules, the MAFN can effectively integrate multimodal information at different scales, improve segmentation accuracy, and achieve good performance in the RRSIS task.

To evaluate the effectiveness of the proposed method, we conduct extensive experiments on the latest remote sensing dataset, RRSIS-D. Experiments demonstrate the superiority of our method: MAFN achieves competitive performance, outperforming previous state-of-the-art methods by 0.93\% and 0.56\% in terms of mean IoU on the validation and test sets, respectively.

The main contributions of our letter are as follows:

\begin{list}{}{}
\item{1)  We design the CFM, a multimodal fusion encoder based on AN-SwinT, leveraging text features generated by BERT \cite{bert} and aligns them with visual features through a correlation fusion algorithm.}
\item{2)  We propose MSRC, a multi-scale convolution operation that effectively enhances segmentation accuracy by refining edge information through multi-scale processing.}
\end{list}

\begin{figure*}[!t]
	\centering
	\includegraphics[width=0.8\textwidth]{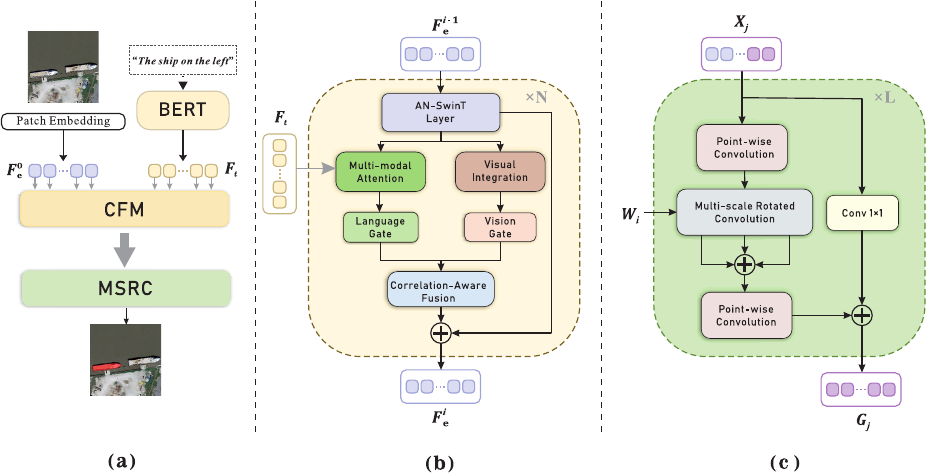}
	\caption{ (a) Overall architecture of the proposed MAFN framework. The framework main consists of CFM and MSRC components. (b) Illustration of our CFM. It contains a Visual Integration Branch, a Multi-modal Attention Branch and a correlation-aware fusion module to achieve pixel-wise fusion of the two modalities. (c) Illustration of our proposed MSRC module. MSRC utilizes different kernel sizes to capture refined multi-scale context features.}
	\label{fig1}
\end{figure*}

\section{Methodology}
\subsection{Overview}
We introduce MAFN and the overall architecture of the proposed method is depicted in Fig. \ref{fig1}(a). Given an image $\textit{I}\in\mathbb{R}^{C\times H\times W}$ and an language expression $E=\left\{p_i\right\}_{i=1}^M$ as the query, where $H$, $W$ represent the height and width of input image, and $M$ is the length of the expression. The input language expression is feed into a text encoder generated by the pre-trained BERT and generate the spatial feature $F_t\in\mathbb{R}^{C\times M}$. Subsequently, the CFM module aligning multi-scale semantic features with visual features to generates $F_e^i$, where $i$ represents the features from the $i$-th layer. To perceive features at different scales, we apply scale transformation to $F_e^i$, generating a new sequence of features $X_j$. Finally, we design an MSRC to perform inference on $X_j$ at different scales to generate segmentation masks.

\subsection{Correlation Fusion Module}

In this part, we describe the design of CFM. As shown in Fig. \ref{fig1}(b), we employ CFM to explore spatial features at different scales and promote fine-grained alignment between linguistic and visual features. Inspired by \cite{geminifusion}, we introduce the minimal amount of noise into the AN-SwinT layer, using adaptive noise to enhance self-attention. We obtain the text features $F_t$ from the pre-trained BERT. The output process of CFM can be described as:

\begin{equation}
	\label{mcf}
	F_e^i = \mathrm{CFM}(F_e^{i-1}, F_t) 
\end{equation}

\noindent  where $i \in [1, . . . , N]$ and $F_e^i$ represents the feature outputted by $\mathrm{CFM}$. Specifically, $F_e^0$ is extracted from a linear patch embedding layer through the input image $\textit{I}$. At each layer, we downsample the input $F_e^{i-1}$. Inspired by RMSIN, the downsampled features are then feed into two branches to enhance visual priors and align multimodal features, respectively.

\subsubsection{Adaptive Noisy Swin Transformer Block}

Since both $K$ and $V$ are derived from the same input, leading the inherent bias in the self-correlation of attention scores. We introduce an attention discriminator based on Swin Block, and add a small amount of noise to reduce the influence of inherent bias and enhance the self-attention capability. The module can be expressed mathematically as follows:

\begin{align}
	\mathrm{Q}&=F_e^{i-1} \mathrm{~W}_q^i \\
	\mathrm{~K}&=\left[\left(\delta_{i}^{k}+F_e^{i-1}\right) \mathrm{W}_k^i, F_e^{i-1} \mathrm{DA}\left(F_e^{i-1}\right) \mathrm{W}_k^i\right] \\
	\mathrm{V}&=\left[\left(\delta_{i}^{v}+F_e^{i-1}\right) \mathrm{W}_v^i, F_e^{i-1} \mathrm{~W}_v^i\right] \\
	V_e^i&=\operatorname{Att}(\mathrm{Q}, \mathrm{K}, \mathrm{V})+F_e^{i-1}
\end{align}

\noindent where $\mathrm{~W}_q^i$ $\mathrm{~W}_k^i$ and $\mathrm{~W}_v^i$ are weight matrix. $\mathrm{DA}(\cdot$) refers to the discriminator, a two-layer MLP followed by a softmax function that assigns attention scores ranging from 0 to 1 to optimize the original keys. We introduce learnable noise, $\delta_i^k$ and $\delta_i^v$, into the key-value pairs at each layer, which are updated during training. This specific noises can enhance the generalization performance of the modality and achieve balance between modalities. $\mathrm{Att}$ is the self-attention layer. We can directly load the checkpoint of Swin Transformer pre-trained without requiring retraining, reducing unnecessary training overhead.

\begin{figure}[!t]
	\centering
	\includegraphics[width=3in]{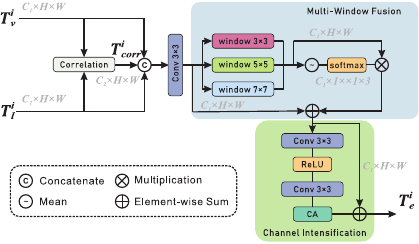}
	\caption{The correlation-aware fusion module.}\label{fig2}
\end{figure}

\subsubsection{Visual Integration Branch}

For features at different scales, their importance is not the same. We treat the features $V_e^i$ go through convolution operations of diverse convolution kernel settings to obtain the receptive field weights at different scales. The visual integration representation can be enhanced by different receptive field weights, which can be formulated as:

\begin{equation}
	G_v^i=\operatorname{sig}\left(\sum_{j=1}^J\left(\frac{1}{C} \sum_{c=1}^C k_j^i * V_e^i\right)\right) \otimes V_e^i
\end{equation}

\noindent where the $k_j^i$ means the $j$-th branch of convolution operation and the $\mathrm{sig}$ is the sigmoid function. In addition, to control the information flowing into the correlation-aware fusion module, the output is regulated by a vision gate, which learns the current weight maps and re-scales the size of the $G_v^i$. The computation process of the gate can be represented as:

\begin{equation}
	T_v^i=\vartheta_i\left(G_v^i\right) \odot G_v^i
\end{equation}

\noindent where $\odot$ denotes element-wise multiplication, and $\vartheta_i$ is a sequential neural network block that performs adaptive modulation of the visual features, with two linear transformations and two non-linear activation functions.

\subsubsection{Multi-modal Attention Branch}
Contextual information is key in RRSIS. We align text features with visual grids to achieve multi-modal attention, ensuring the effective fusion of semantic and spatial features across different modalities. Specifically, input visual features $V_e^i$ generated by AN-SwinT and text features $F_t$, using $V_e^i$ as query values and $F_t$ as keys and values to generate multi-modal features:

\begin{equation}
	G_l^i=\mathrm{Proj}(\mathrm{Att}(V_e^iW_{eq}^i,F_tW_{lk}^i,F_tW_{lv}^i) \odot V_e^iW_e^i)
\end{equation}

\noindent where $W_{eq}^i$, $W_{lk}^i$, $W_{lv}^i$ and $W_e^i$ are the projection matrices. The obtained feature is generated through a 1×1 convolution, indicated as $\mathrm{Proj}(\cdot)$. Similar to the visual integration branch, we add the same operations $\phi_i$ as $\vartheta_i$ at the end to further adjust the language information:

\begin{equation}
	T_l^i=\phi_i\left(G_l^i\right) \odot G_l^i
\end{equation}

\begin{figure}[!t]
	\centering
	\includegraphics[width=3in]{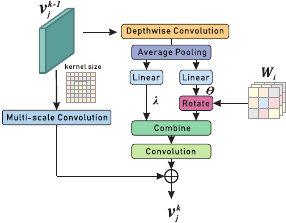}
	\caption{Pipeline of the multi-scale rotated convolution.}\label{fig3}
\end{figure}

\subsubsection{Correlation-aware Fusion Module}

We further enhance the interaction between the two branches and achieve fine-grained fusion between the modalities as shown in Fig. \ref{fig2}. Inspired by optical flow estimation \cite{lite}, we perform correlation volume calculations on input $T_v^i\in\mathbb{R}^{C_1\times H\times W}$ and $T_l^i\in\mathbb{R}^{C_1\times H\times W}$ to compare their similarity and generate $T_{corr}^i\in\mathbb{R}^{C_2\times H\times W}$ by  position offset, where $C_1$, $C_2$ respectively represent number of feature channels and offsets:

\begin{equation}
	T_{corr}^i=\left[\frac{1}{C_1} \sum_{c=1}^{C_1} T_v^i(c, h, w) \cdot T_l^i(c, h+\Delta x, w+\Delta y)\right]_{\Delta x, \Delta y}
\end{equation}

\noindent where $\Delta x$ and $\Delta y$ represent horizontal and vertical offsets. Outputs from correlation stages, $T_v^i$, $T_l^i$ and $T_{corr}^i$ are concatenated and fed into the multi-window fusion for refining. Multi-window fusion begins with a smooth transition of 3×3 convolution, extracting multi-scale features by employing different window sizes e.g., 3×3, 5×5, 7×7. Afterwards, we calculate the weighted coefficients to fuse these features, followed by a residual connection.

In channel intensification, we process feature map by the two convolution layers and the ReLU activation function. Subsequently, the feature map is intensified by utilizing the channel attention mechanism (CA) to enrich the channel details of the features. Similar to the operation of multi-window fusion, the output $T_e^i$ is connected by a residual block. Consequently, the overall output features of CFM at $i$-layer can be illustrated as:

\begin{equation}
	F_e^i=V_e^i+T_e^i
\end{equation}

\subsection{Multi-scale Refinement Convolution}
In this section, we introduce a novel MSRC as shown in Fig. \ref{fig1}(c), which generate more refined segmentation for aerial images with spatial complexity and multiple object orients. To effectively segment hierarchical referent feature, we extend the adjacent layer features obtained from CFM to the same dimension by bilinear interpolation and concatenate them to generate $X_j$. The overall top-down process can be concluded as follows:
 
\begin{align}
	G_L &= F_e^N \\
	X_j &= \mathrm{Concat}(\mathrm{Interpolate}(G_j,\text{size}(F_e^j),F_e^j)) \\
	G_{j-1} &= \mathrm{Dec}(\mathrm{MSRC}(X_j,W_i)) 
\end{align}

\noindent here, $j \in [1, . . . , L]$ and $\mathrm{Dec}(\cdot)$ refers to a 3×3 convolution followed by a batch normalization layer and a ReLU activation function.

 \begin{table*}[!t]
 	\begin{center}
 		\caption{ Comparison with state-of-the-art methods on the RRSIS-D dataset. R-101 \cite{res} and Swin-B represent ResNet-101 and base Swin Transfomer  models, respectively. The best result is highlighted in bold.}
 		\begin{tabularx}{\textwidth}{l|c|c|X|X|X|X|X|X|X|X|X|X|X|X|X|X}
 			\Xhline{0.8pt} \noalign{\vskip 2pt}
 			\multirow{2}{*}{Method}&\multirow{2}{*}{\makecell{Visual\\Encoder}}&\multirow{2}{*}{\makecell{Text\\Encoder}} &\multicolumn{2}{c|}{P@0.5} &\multicolumn{2}{c|}{P@0.6} &\multicolumn{2}{c|}{P@0.7} 
 			&\multicolumn{2}{c|}{P@0.8} &\multicolumn{2}{c|}{P@0.9} &\multicolumn{2}{c|}{oIoU} &\multicolumn{2}{@{}c@{}}{mIoU} \\ \cline{4-17} 
 			& &&Val&Test&Val&Test&Val&Test&Val&Test&Val&Test&Val&Test&Val&Test\\ \noalign{\vskip 2pt}
 			\hline \noalign{\vskip 2pt}
% 			RRN &R-101 &LSTM &51.09 &51.07 &42.47 &42.11 &33.04 &32.77 &20.80 &21.57 &6.14 &6.37 &66.53 &66.43 &46.06 &45.64 \\
% 			CSMA&R-101 &None &55.68 &55.32 &48.04 &46.45 &38.27 &37.43 &26.55 &25.39 &9.02 &8.15 &69.68 &69.39 &48.85 &48.54 \\
% 			LSCM&R-101 &LSTM &57.12 &56.02 &48.04 &46.25 &37.87 &37.70 &26.37 &25.28 &7.93 &8.27 &69.28 &69.05 &50.36 &49.92 \\
 			CMPC&R-101 &LSTM &57.93 &55.83 &48.85 &47.40 &38.50 &36.94 &25.28 &25.45 &9.31 &9.19 &70.15 &69.22 &50.41 &49.24 \\
 			BRINet&R-101&LSTM&58.79 &56.90 &49.54 &48.77 &39.65 &39.12 &28.21 &27.03 &9.19 &8.73 &70.73 &69.88 &51.14 &49.65 \\
 			CMPC$+$&R-101 &LSTM &59.19 &57.65 &49.36 &47.51 &38.67 &36.97 &25.91 &24.33 &8.16 &7.78 &70.14 &68.64 &51.41 &50.24 \\
 			LGCE&Swin-B &BERT &68.10 &67.65 &60.52 &61.53 &52.24 &51.45 &42.24 &39.62 &23.85 &23.33 &76.68 &76.34 &60.16 &59.37\\
 			LAVT&Swin-B &BERT &69.54 &69.52 &63.51 &63.63 &53.16 &53.29 &43.97 &41.60 &24.25 &24.94 &77.59 &77.19 &61.46 &61.04\\
 			RMSIN&Swin-B&BERT &74.66 &74.26 &68.22 &67.25 &57.41 &55.93 &\textbf{45.29} &42.55 &24.43 &\textbf{24.53} &78.27 &\textbf{77.79} &65.10 &64.20\\
 			\noalign{\vskip 2pt} \hline
 			%%	\rowcolor{gray!25}
 			MAFN(ours)&Swin-B &BERT & \textbf{76.32} & \textbf{75.27} & \textbf{69.31} & \textbf{68.14} & \textbf{58.33} &\textbf{56.79}&44.54&\textbf{43.49}&\textbf{24.71}&23.76&\textbf{78.33}&77.41&\textbf{66.03}&\textbf{64.76}\\
 			\Xhline{0.8pt}
 		\end{tabularx}
 		\footnotetext{ The best score is highlighted in bold.}
 		\label{table1}
 	\end{center}
 \end{table*}

\subsubsection{Multi-scale Rotated Convolution}

Specifically, our MSRC first uses point-wise convolution to expand the channel dimensions, and then inputs into the $k$-layer multi-scale rotated convolution, where $k\in[1, . . . , K]$. The overall pipeline is illustrated in Fig. \ref{fig3}, we introduce a multi-scale convolution branch on the basis of adaptive rotated convolution (ARC) in \cite{adaptive} and utilize multi-scale rotated convolution as a hierarchical structure. In ARC branch, the input feature $V_j^{k-1}$ is feed into a depthwise convolution, followed by average pooling to predict a set of possible angles $\theta = [\theta_1, \theta_2, \cdots, \theta_n]$ and weights $\lambda=[\lambda_1, \lambda_2, \cdots,  \lambda_n]$. We initialize $n$ learnable convolution kernels $W_i$ to perform adaptive rotation according to possible prediction angles:

\begin{equation}
W_i' = \mathrm{Rotate}(W_i; \theta_i), \, i = 1, 2, \dots, n,
\end{equation}

\noindent where $\mathrm{Rotate}(\cdot)$ is a rotation operation, which performs bilinear interpolation on the original convolution kernel span space and resamples the rotated convolution kernel. The weights are integrated into the convolution by combining with the convolution kernels, and finally sum the output feature maps in an element-wise manner: 

\begin{equation}
R_j^{k-1} = V_j^{k-1} * \sum_{i=1}^{n} \lambda_i W_i'
\end{equation}

\noindent here, $*$ is the convolution operation. In the multi-scale convolution branch, we design a multi-scale convolution and treat the multi-scale rotated convolution as a hierarchical structure. In each layer, the multi-scale convolution adopts different kernel sizes. Finally, we combine the two branch outputs to preserve the feature maps for each layer.

\begin{equation}
	V_j^k = V_j^{k-1} * S_k+R_j^{k-1}
\end{equation}

\noindent where $S_k$ and $V_j^k$ are the $k$-th different convolution kernel and output features of multi-scale rotated convolution. We sum all features and use another point-by-point convolution to convert back to the original channel. As detailed in the ablation studies in Table \ref{table3}, we choose three-layer convolution kernels [1,3,5] in MSRC, which can improve the boundary details at different scales to a certain extent.

\section{Experiment}
\subsection{Implementation Details}
\subsubsection{Experiment Settings}

We implement our method in PyTorch based on the transformer model. The visual encoder layer is initialized with pre-trained parameters on ImageNet22K from Swin Transformer, with a window size set to 12. The language encoder layer employs the base BERT model from the HuggingFace library, initialized with the official pre-trained weights. We train MAFN with optimizer AdamW and an initial learning rate of 3e-5. The number of layers of the CFM and MSRC are set to 4 and 3 layers respectively. The model was trained for 40 epochs with a weight decay of 0.01. We conduct our experiments with a batch size of 8 on two NVIDIA A40 GPUs.

\begin{figure}[!t]
	\centering
	\includegraphics[width=1\linewidth]{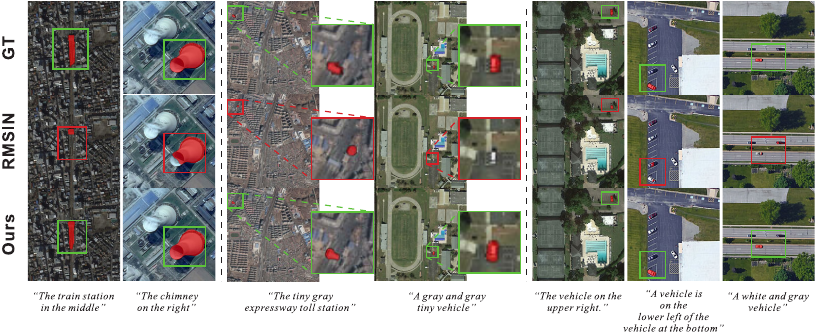}
	\caption{Qualitative comparisons between MAFN and the previous SOTA RMSIN in various scales and directions. The left part illustrates predictions of large-scale objects, while the middle part provides predictions for tiny-scale targets in noisy environments. The right part displays predictions for object at different angles. Please zoom in for better visualization.}\label{fig4}
\end{figure}

\subsubsection{Datasets and Metrics}
We perform the experimental analysis on the RRSIS-D dataset designed specifically for the RRSIS task. The RRSIS-D dataset includes 17,402 remote sensing images with a fixed dimension of 800 $\times$ 800 pixels. The training, validation, and test sets consist of 12,181, 1,740 and 3,481 images, respectively. All images are accompanied by corresponding pixel-level masks annotations and reference expressions. It contains a total of 20 semantic label categories and 7 attributes. Following previous work, we adopt overall intersection-over-union (oIoU), mean intersection-over-union (mIoU), and precision at thresholds ranging from 0.5 to 0.9 as evaluation metrics.

\subsection{Comparison With State-of-the-Art}
We conduct experiments to compare the performance of our proposed MAFN with the state-of-the-art referring image segmentation methods. The validation and test results on the RRSIS-D dataset are shown in Table \ref{table1}. For a fair comparison, we adopt experimental results from original paper RMSIN. The performance of MAFN almost outperforms the all previous methods on both subsets. In specific, our model has further improvement over RMSIN with 66.03\% and 64.76\% in mIoU on the validation and test subsets respectively. It is worth noting that MAFN can still maintain satisfactory performance when detecting small or different rotated objects, with higher performance improvement below the IoU threshold of 0.7. Fig. \ref{fig4} illustrates visualization examples of the results obtained at various scales compared with RMSIN. It can be observed that MAFN can more accurately segment ground objects with less noise.
\begin{table}[h]
	\caption{ Comparison of variants
		of our baseline }
	\begin{tabularx}{\columnwidth}{l|XXXXX}
		\Xhline{0.8pt}%
		\noalign{\vskip 2pt}
		Options & P@0.5 & P@0.7 & P@0.9 & oIoU & mIoU\\ \noalign{\vskip 2pt}
		\hline
		\noalign{\vskip 2pt}
		default &75.23 &55.4 &24.2 &77.77 &64.8  \\ \noalign{\vskip 2pt}
		\hline
		\noalign{\vskip 2pt}
		+ Correlation-Aware Fusion &75.29 &56.26 &23.45 &77.46 &64.92\\
		+ Adaptive Noisy&75.75 &56.95 &24.2 &77.88 &65.14 \\
		+ MSRC &\textbf{76.32} &\textbf{58.33} &\textbf{24.71} &\textbf{78.33} &\textbf{66.03} \\ \noalign{\vskip 2pt}
		\Xhline{0.8pt}
	\end{tabularx}
	\label{table2}
	
\end{table}

\begin{table}[h]
	\caption{Effect of different kernels in multi-scale rotated convolutions of MSRC}
	\begin{tabularx}{\columnwidth}{l|XXXXX}
		\Xhline{0.8pt}%
		\noalign{\vskip 2pt}
		kernels & P@0.5 & P@0.7 & P@0.9 & oIoU & mIoU\\ \noalign{\vskip 2pt}
		\hline
		\noalign{\vskip 2pt}
		\textnormal{[1]}      & 74.2  & 56.38 & 23.79 & 77.07 & 64.94 \\ \noalign{\vskip 2pt}
		\textnormal{[1,3]} & 74.83 & 55.63 & 24.08 & 77.79 & 64.54\\
		\textnormal{[1,3,5]}& 76.32 & \textbf{58.33} & \textbf{24.71} & 78.33 & \textbf{66.03}  \\
		\textnormal{[1,3,5,7]}  & \textbf{76.44} & 57.76 & 24.31 & \textbf{78.53} & 65.75 \\ \noalign{\vskip 2pt}
		\Xhline{0.8pt}
	\end{tabularx}
	\label{table3}
	
\end{table}

%\begin{table}[h]
%\caption{Effect of different kernels in multi-scale rotated convolutions of MSRC}
%\begin{tabularx}{\columnwidth}{l|XXXXX}
%	\Xhline{0.8pt}
%	kernels & P@0.5 & P@0.7 & P@0.9 & oIoU & mIoU \\ 
%	\Xhline{0.8pt}
%	[1]     & 74.2  & 56.38 & 23.79 & 77.07 & 64.94 \\ 
%	[1,3]   & 74.83 & 55.63 & 24.08 & 77.79 & 64.54 \\ 
%	[1,3,5] & 76.32 & 58.33 & 24.71 & 78.33 & 66.03 \\ 
%	[1,3,5,7] & 76.44 & 57.76 & 24.31 & 78.53 & 65.75 \\
%	\Xhline{0.8pt}
%\end{tabularx}
%\end{table}

\begin{figure}[!t]
	\centering
	\includegraphics[width=\linewidth]{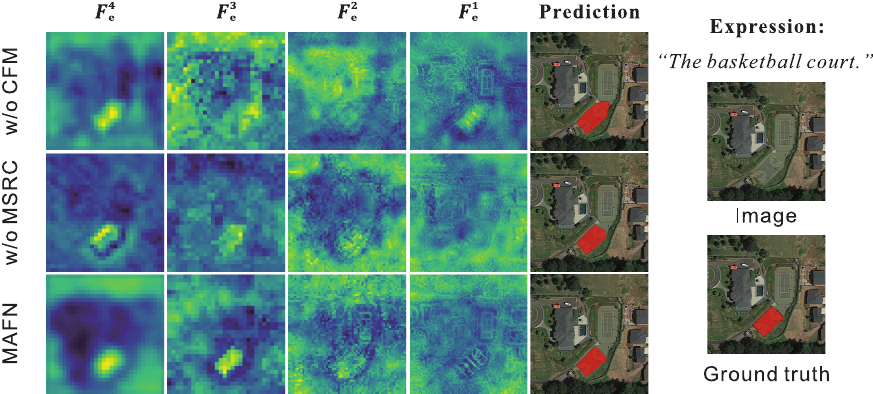}
	\caption{Visualization of predictions and feature maps from validation set, the i-th column represents the output feature map from the i-th stage of the CFM. Each row
		shows the results of stepwise addition of modules.}\label{fig5}
\end{figure}

\subsection{Ablation study}
\subsubsection{Comparison with the baseline}
We have conducted a series of ablation experiments on the validation subset of RRSIS-D to assess the efficacy of the various components within our proposed network and the results are shown in Table \ref{table2}. We adapt the design of RMSIN and introduce a default model to prove the validity of each module by cumulative modules. We observe adaptive noise has discernible improvements in segmentation precision without additional training burden. Furthermore, MSRC demonstrate significant performance promotion at lower IoU thresholds, expositing an enhancement of over 2.93\% in metrics.

\subsubsection{Effect of multi-scale kernels in MSRC}
We further conduct a set of experiments to investigate the the effect of different multi-scale kernels used for convolutions in MSRC. As demonstrated in Table \ref{table3}, with the increase in the number of multi-scale convolution kernel, the performance of the model on the RRSIS-D improves. If we add additional larger 7×7 kernel, the performance improvement is not noticeable. Based on these experimental results, we choose three-layer convolution kernels [1,3,5] in all experiments.

\subsection{Visual Analysis}
In Fig. \ref{fig5}, we visualize the feature maps of MAFN during training without CFM or MSRC. From the feature maps, it is obvious that MAFN accurately locates semantic features based on the expression and effectively captures boundary information. Comparing the predicted masks between the three models, we can observe that CFM keenly focuses on relevant target regions in the deep layer, while MSRC refines edge features, resulting in improved segmentation accuracy. 
\section{Conclusion}
In this letter, we propose MAFN, a novel network structure designed for perceptual fusion of multimodal features across complex scales. The framework incorporates adaptive noise into transformer without incurring additional training costs, which then combines visual integration branch and multi-modal attention branch for relational alignment. To further effectively address deflection orientations in complex aerial images, we devise multi-scale convolution via adopting different kernels. Extensive experiments demonstrate its advantage over the state of the art in the RRSIS task.

\bibliographystyle{IEEEtran}
\bibliography{IEEEabrv,main}

\begin{thebibliography}{10}
\providecommand{\url}[1]{#1}
\csname url@samestyle\endcsname
\providecommand{\newblock}{\relax}
\providecommand{\bibinfo}[2]{#2}
\providecommand{\BIBentrySTDinterwordspacing}{\spaceskip=0pt\relax}
\providecommand{\BIBentryALTinterwordstretchfactor}{4}
\providecommand{\BIBentryALTinterwordspacing}{\spaceskip=\fontdimen2\font plus
\BIBentryALTinterwordstretchfactor\fontdimen3\font minus \fontdimen4\font\relax}
\providecommand{\BIBforeignlanguage}[2]{{%
\expandafter\ifx\csname l@#1\endcsname\relax
\typeout{** WARNING: IEEEtran.bst: No hyphenation pattern has been}%
\typeout{** loaded for the language `#1'. Using the pattern for}%
\typeout{** the default language instead.}%
\else
\language=\csname l@#1\endcsname
\fi
#2}}
\providecommand{\BIBdecl}{\relax}
\BIBdecl

\bibitem{rrsis}
Z.~Yuan, L.~Mou, Y.~Hua, and X.~X. Zhu, ``Rrsis: Referring remote sensing image segmentation,'' \emph{IEEE Transactions on Geoscience and Remote Sensing}, vol.~62, pp. 1--12, 2024.

\bibitem{transvg}
J.~Deng, Z.~Yang, T.~Chen, W.~Zhou, and H.~Li, ``Transvg: End-to-end visual grounding with transformers,'' in \emph{Proceedings of the IEEE/CVF International Conference on Computer Vision}, 2021, pp. 1769--1779.

\bibitem{rsvg}
Y.~Zhan, Z.~Xiong, and Y.~Yuan, ``Rsvg: Exploring data and models for visual grounding on remote sensing data,'' \emph{IEEE Transactions on Geoscience and Remote Sensing}, vol.~61, pp. 1--13, 2023.

\bibitem{lavt}
Z.~Yang, J.~Wang, Y.~Tang, K.~Chen, H.~Zhao, and P.~H. Torr, ``Lavt: Language-aware vision transformer for referring image segmentation,'' in \emph{Proceedings of the IEEE/CVF Conference on Computer Vision and Pattern Recognition}, 2022, pp. 18\,155--18\,165.

\bibitem{RMSIN}
S.~Liu, Y.~Ma, X.~Zhang, H.~Wang, J.~Ji, X.~Sun, and R.~Ji, ``Rotated multi-scale interaction network for referring remote sensing image segmentation,'' in \emph{Proceedings of the IEEE/CVF Conference on Computer Vision and Pattern Recognition (CVPR)}, June 2024, pp. 26\,658--26\,668.

\bibitem{swin}
Z.~Liu, Y.~Lin, Y.~Cao, H.~Hu, Y.~Wei, Z.~Zhang, S.~Lin, and B.~Guo, ``Swin transformer: Hierarchical vision transformer using shifted windows,'' in \emph{Proceedings of the IEEE/CVF international conference on computer vision}, 2021, pp. 10\,012--10\,022.

\bibitem{bert}
J.~Devlin, ``Bert: Pre-training of deep bidirectional transformers for language understanding,'' \emph{arXiv preprint arXiv:1810.04805}, 2018.

\bibitem{geminifusion}
D.~Jia, J.~Guo, K.~Han, H.~Wu, C.~Zhang, C.~Xu, and X.~Chen, ``Geminifusion: Efficient pixel-wise multimodal fusion for vision transformer,'' \emph{arXiv preprint arXiv:2406.01210}, 2024.

\bibitem{lite}
T.-W. Hui, X.-A. F. N. F. R. R. S. I.~S. Tang, and C.~C. Loy, ``Liteflownet: A lightweight convolutional neural network for optical flow estimation,'' in \emph{Proceedings of the IEEE conference on computer vision and pattern recognition}, 2018, pp. 8981--8989.

\bibitem{res}
K.~He, X.~Zhang, S.~Ren, and J.~Sun, ``Deep residual learning for image recognition,'' in \emph{Proceedings of the IEEE conference on computer vision and pattern recognition}, 2016, pp. 770--778.

\bibitem{adaptive}
Y.~Pu, Y.~Wang, Z.~Xia, Y.~Han, Y.~Wang, W.~Gan, Z.~Wang, S.~Song, and G.~Huang, ``Adaptive rotated convolution for rotated object detection,'' in \emph{Proceedings of the IEEE/CVF International Conference on Computer Vision}, 2023, pp. 6589--6600.

\end{thebibliography}

\end{document}